\documentclass[letterpaper, 10 pt, conference]{ieeeconf}  % Comment this line out if you need a4paper

\IEEEoverridecommandlockouts                              % This command is only needed if 
\usepackage{amsmath}
\usepackage{amsfonts}
\usepackage{bbm}
\usepackage{array}
\usepackage{multirow}
\usepackage{graphicx}
\usepackage{caption}
\usepackage{subcaption}
\usepackage{fancyhdr}

\newcommand{\bvec}[1]{\mathbf{#1}}

\title{\LARGE \bf
Intent-Aware Autonomous Driving:\\
A Case Study on Highway Merging Scenarios
}

\author{Nishtha Mahajan$^{1}$ and Qi Zhang$^{2}$% <-this % stops a space
\thanks{$^{1}$Nishtha Mahajan is an independent researcher, work done at the University of South Carolina
        {\tt\small nmahaja@alumni.purdue.edu}}%
\thanks{$^{2}$Qi Zhang is with The Artificial Intelligence Institute, University of South Carolina,
        Columbia, SC 29208, USA
        {\tt\small qz5@cse.sc.edu}}%
\thanks{\hrule\vspace{2mm}}%
\thanks{MAD-Games: Multi-Agent Dynamic Games Workshop, IROS 2023}%
}

\begin{document}

\maketitle
% \fancyfoot[C]{MADGames Workshop, IROS 2023}
\thispagestyle{empty}
\pagestyle{empty}

%%%%%%%%%%%%%%%%%%%%%%%%%%%%%%%%%%%%%%%%%%%%%%%%%%%%%%%%%%%%%%%%%%%%%%%%%%%%%%%%
\begin{abstract}

In this work, we use the communication of intent as a means to facilitate cooperation between autonomous vehicle agents. Generally speaking, intents can be any reliable information about its future behavior that a vehicle communicates with another vehicle. We implement this as an intent-sharing task atop the merging environment in the simulator of highway-env, which provides a collection of environments for learning decision-making strategies for autonomous vehicles. Under a simple setting between two agents, we carefully investigate how intent-sharing can aid the receiving vehicle in adjusting its behavior in highway merging scenarios.

\end{abstract}

%%%%%%%%%%%%%%%%%%%%%%%%%%%%%%%%%%%%%%%%%%%%%%%%%%%%%%%%%%%%%%%%%%%%%%%%%%%%%%%%
%-----------------------
\section{Introduction} 
% AV-AV communication
Autonomous vehicles (AVs) hold the promise of improving the driving of not only individual vehicles but also those with whom they interact. Of special interest is AVs’ ability to communicate with both infrastructure and other AVs that allows them to act in ways that human drivers cannot. 
% For instance, we know from game theory that cooperation is always better than each driver pursuing their own interests. However, human drivers can fail to commit to cooperation when other drivers' intentions are unknown. AVs can overcome this limitation since they can communicate with each other.
% Our focus: AV-AV intent communication
This work seeks to leverage the communication capability of AVs to facilitate cooperation among them. Specifically, we focus on investigating how by sharing some information about their future behavior, which we formally call \textit{intent}, AVs can influence each other’s behavior positively. 

\subsection{What is Intent-Sharing among AVs and Why?}
% Given the value of cooperating, several past works have looked at inducing cooperative behavior in autonomous vehicles both with and without communication. While \citet{toghi2022} use the concept of social value orientation to elicit altruism-based cooperation among AVs, 

As per SAE (Society of Automotive Engineers) Standard on Cooperative Driving Automation \cite{on2021taxonomy}, intent-sharing is a type of cooperative driving automation facilitated by machine to machine communication. Here, an intent is defined to be some information about the sending entity's future actions that it shares to aid the decision-making of the receiving entity. 
Intent-sharing is different from status-sharing in the sense that the status only contains information about what the sending entity observes at present, while the intent contains information about the future.
The Standard also notes that the receiving entity does not have to accept the information from the sender. The sender executes its planned trajectory irrespective of how the receiver responds. It is different from other forms of cooperation where interaction between entities may be required to arrive at an agreed cooperation plan.
% Readers can refer to SAE's technical report J3216 \cite{on2021taxonomy} for more details.

% Why: for the receiver and better coorperative driving behavior
Although the receiver might have limited influence on the sender's intent choice, the receiver gains better prediction of the sender's prolonged future trajectory and therefore, by planning accordingly, can potentially achieve cooperative driving behavior that is beneficial for both entities.
% The example of highway merging
As an example that we will investigate carefully in this work, consider a scenario where a vehicle aims to merge into a highway from the entrance ramp, where there is another incoming vehicle on the highway that can potentially block the entrance.
The incoming vehicle can either stay at the same speed so that the merging vehicle has to yield, or change its lane or speed or both so that the merging vehicle can safely enter without hesitation.
Either way, for better cooperative driving, it is crucial for the merging vehicle to know the intent of the incoming vehicle.

\subsection{How to Implement Intent-Aware Autonomous Driving?}
It is non-trivial to implement the aforementioned concept, what we call intent-aware autonomous driving (AD), for effective cooperative driving, as the following three questions have to be addressed.
Q1: How to computationally define and represent intents for AVs?
Q2: Given a specific intent, how does its sender comply with the intent (i.e.,  plan and execute a trajectory consistent with the intent), and how does its receiver utilize the intent (e.g., whether or not to wait at the entrance ramp)?
Q3: How to choose an intent to induce the optimal cooperative driving behavior (e.g., to yield to the merging vehicle, or not)?

Here, Q1 aims to provide a computational framework as the foundation for intent-aware AD, whereas Q2 and Q3 are coupled together to provide an algorithmic solution under the framework.

\subsection{Our Contributions}
In this work, we view AVs as autonomous decision-making agents and formulate a decision-theoretic framework for intent-aware AD (Section \ref{sec:prob_set}).
As a case study, we implement intent-aware AD for highway merging scenarios in the simulator of highway-env \cite{highway-env} (Section \ref{sec:implementation}).
Our implementation answers Q1 and Q2:
we design a representation of intents for the highway-env scenarios that facilities the intent-sender to comply with its intent, while the intent-receiver adopts a reinforcement learning (RL) approach to utilize the intent. 
Our results demonstrate that our implementations achieve more reliable and more efficient driving behavior (Section \ref{sec:results}).
Finally, we leave Q3 as our future work, along with several other future directions that can be built on our intent-aware AD framework (Section \ref{sec:conclusion}).

% Our main contributions are:
% \begin{itemize}
% \item We introduce a new intent-sharing task for autonomous driving. Our implementation is built upon the framework of highway-env \cite{highway-env}, which is a collection of environments for learning decision-making for autonomous driving. 
% \item Using a simple experimental setup, we investigate the potential of intent-sharing to influence the behavior of other AVs.  
% \end{itemize}

% The rest of the paper is organized as follows: We describe intent-sharing and discuss related works on intent in Section \ref{sec:background}. After that, we present our problem setting and formally introduce intent in Section \ref{sec:prob_set}. Thereafter, we describe our implementation in Section \ref{sec:implementation} and the experimental setup we use in Section \ref{sec:experiments}. We then present our results and discussion in Section \ref{sec:results} followed by conclusion and a discussion on our future work in Section \ref{sec:conclusion}. 

%-----------------------
%\section{Related Works}

% Different forms of cooperative behavior

% Different types of decentralized cooperative behavior

% What we consider

%-----------------------
\section{Related Work}
\label{sec:Related Work}

% Sharing vs inferring intent
% When using communication to encourage cooperation between AVs, status-sharing has been extensively used. 
% Status-sharing has been extensively used as a means to encourage cooperation among AVs \cite{cui2022coopernaut, toghi2022social}. While \citet{cui2022coopernaut} look at cooperative perception, \citet{toghi2022social} use the concept of social value orientation to elicit cooperation by controlling the degree of altruism of AVs.

% There have also been attempts to elicit cooperation implicitly through rewards. For instance,  
\subsection{Intent and Intent-Sharing}
Past research has focused both on \textit{inferring} the intent of other agents \cite{bandyopadhyay2013intention, qi2018intent} and \textit{explicitly communicating} an agent's own intent \cite{wu2021spatial, kim2021communication}.
While inferred intent is usually an agent’s goal or plan recognized based on the history of its activity, shared intent is future-directed that the agent is yet to reveal through its actions. 
For instance, Qi and Zhu \cite{qi2018intent} use agents' goal locations to represent intents, which are inferred based on an observation history. 
On the other hand, 
% Wu et al. \cite{wu2021spatial} consider an agent's last chosen action as its intention and communicate waypoints (i.e. $(x,y)$ coordinates) of its intended path to other agents.
Kim et al. \cite{kim2021communication} communicate intentions between agents by generating each agent's imagined trajectory and encoding them as intention messages using an attention mechanism.
% Their numerical results show improved performance as compared to approaches when only partial observations are shared among agents. 
We are interested in the latter case of explicit communication, which, as we will show, allows the intent-receiver agent to adapt its behavior to the communicated intent right from the start. 
% to utilize the communicated intent before it can be inferred.

In the AD space, intent-sharing has been shown to allow AVs to undertake less conservative but safe driving maneuvers \cite{wang2022multi}. 
To this end, Wang et al. use reachability theory and conflict charts to demonstrate the potential of sharing velocity and acceleration bounds over a given time horizon to prevent conflicts during lane changes.
% Wang et al. \cite{wang2022multi} use reachability theory and conflict charts to demonstrate that intent-sharing has the potential to allow AVs to undertake less conservative but safe driving maneuvers. They use velocity and acceleration bounds over a given time horizon as their representation of intent.
Beyond AV-AV settings, Matthews et al. \cite{matthews2017intent} choose between the decisions to display and not display the intention of an AV to pedestrians, and show both through in-field experiments and simulations that intent-communication has a positive impact on the behavior of the receiving entity. 
% Beyond AV-AV settings, Matthews et al. \cite{matthews2017intent} design an intent communication system to convey the intention of an AV to pedestrians. They choose between the decisions to display and not display the intent message, and show both through in-field experiments and simulations that intent-communication has a positive impact on the behavior of the receiving entity. 

% Our work is different from prior work in the sense that the decision in past work on intent-sharing is mostly considered to be the choice between whether to share or not share one's intent. 
% Compliance with a given intent is not necessarily considered.
% On the contrary, we seek to treat the intent as a decision variable that the intent-sender agent has to ensure compliance with.
The decision in intent-sharing is mostly considered to be the choice between whether to share or not share one's intent. 
Compliance with a given intent is not necessarily considered.
On the contrary, we seek to treat intent as a decision variable that the intent-sender agent has to ensure compliance with.

\subsection{RL for Highway Merging}
Highway merging is considered a challenging traffic scenario for AD, owing to the complexity of interactions between vehicles. 
% Given the sequential decision-making that RL facilitates, it has been extensively explored for learning AD behavior.
Several variations of this problem have been considered in literature where behavior of  AVs either on the entrance ramp 
% \cite{triest2020learning, liu2022autonomous, bouton2019cooperation}
and/or the highway 
% \cite{toghi2022social}, 
% or both 
% \cite{tang2019towards} 
is learned using RL.

For instance, 
Tang \cite{tang2019towards} use RL to learn merging policies for AVs using a technique called self-play, wherein they iteratively update the set of AV agents in their training using previously learned policies. While their method achieves a high merge-success rate, it is not collision-free and they state that unobservability of intentions, among other reasons, might be a contributing factor. 
On the other hand, 
Liu et al. \cite{liu2022autonomous} focus on safety and equip their RL method with a motion predictive safety controller to reject unsafe actions that their RL policy might learn for an AV to merge into a stream of simulated human-driven vehicles. 
In contrast,
Triest et al. \cite{triest2020learning} use real human-driving data for merge scenarios and train an RL-based merging vehicle to choose target vehicles it should keep distances with, in order to merge successfully at a fixed merge-point. They find that RL performs better with high-level actions such as theirs as compared to low-level acceleration actions in such cases.

% to leverage or elicit cooperation between AVs.
% RL has also been used for cooperative AD.  
Cooperative AD has also been explored for merging scenarios. 
Bouton et al. \cite{bouton2019cooperation} achieve this by first assigning a cooperation level to human-driven vehicles on the highway. Then they use a belief-state RL method to make the merging AV merge at a fixed point, by inferring the cooperation level of the vehicles it observes.
% while inferring the cooperation level of the vehicles it observes. 
% consider different densities of human-driven vehicles in a highway merging scenario
Contrarily, Toghi et al. \cite{toghi2022social} use multi-agent RL to determine optimal driving behavior of AVs on the highway such that a merging vehicle can successfully merge. They use the concept of social value orientation to 
% elicit cooperation by controlling 
control the degree of altruism of AVs.
% Major trends we see
% Major takeaways here are: 

%-----------------------
\section{Formulation of Intent-Aware AD}  
\label{sec:prob_set}

As a starting point, we focus on the problem of intent-aware AD in the setting of two-AV, one-way intent-sharing between an intent-sender AV (indexed by $2$) and an intent-receiver AV (indexed by $1$).
% Our implementation and experimens focues on such a setting, while we discuss its extension in Section .
We view the AVs as decision-making agents and formulate the problem with the notion of two-agent dynamic game.
% Two-agent dynamic game
Let $\bvec{s}_t$ denote the world state at (discrete) time step $t$, which we assume can be factored as $\bvec{s}_t=[\bvec{s}_t^e, \bvec{s}_t^1, \bvec{s}_t^2]$,
where $\bvec{s}_t^1, \bvec{s}_t^2$ are the two agents' local states and $\bvec{s}_t^e$ is part of the world state external to the two agents.
The world state's (probabilistic) dynamics is denoted as $p(\bvec{s}_{t+1}|\bvec{s}_t, a_t)$, where $a_t=[a_t^1, a_t^2]$ consists of both agents' actions.
The agents are self-interested in the sense that they aim to optimize their respective cumulative reward.
We let $r_t^j=r^j(\bvec{s}_t^j)$ denote the reward at time $t$ for agent $j\in\{1, 2\}$, assuming reward function $r^j(\cdot)$ only depends on agent $j$'s local state.

% Sender's local problem
For agent 2, we denote its intent as $\bvec{i}^2$, which intuitively encodes information about its admissible future trajectories.
Formally, we assume agent 2's control over its local state is independent of agent 1, of which the dynamics is denoted as $p^2(\bvec{s}_{t+1}^2|\bvec{s}_t^e, \bvec{s}_t^2,a_t^2)$, such that it can optimize its trajectory without considering agent 1.
Formally, given an intent $\bvec{i}^2$, agent 2 aims to choose its policy $\pi^2: [\bvec{s}_t^e, \bvec{s}_t^2, \bvec{i}^2] \mapsto a_t^2$ that optimizes its expected cumulative reward while ensuring the future trajectory up to a certain time horizon $H$ is admissible to intent $\bvec{i}^2$:
\begin{align}\label{eq:agent_2_opt}
&\textstyle\max_{\pi^2}~\mathbb{E}\left[\left.\textstyle\sum_{t=0}^{H} r^2_t \right\lvert \bvec{s}_0^e, s^2_0\right]\\
s.t.\quad
&\bvec{s}_{t+1}^2 \sim p^2(\bvec{s}_t^e, \bvec{s}_t^2,a_t^2), \quad t = 0,\ldots,H-1 \nonumber\\
&a_{t}^2 \sim \pi^2(\bvec{s}_t^e, \bvec{s}_t^2, \bvec{i}^2), \qquad t = 0,\ldots,H-1 \nonumber\\ 
& (s^2_{t})_{t=0}^H \in \bvec{i}^2 \nonumber
\end{align}
where in the last constraint we slightly abuse the notation to let $\bvec{i}^2$ denote the admissible set of agent 2's trajectories.

Agent 1 conditions its action selection on the world state, as well as the intent received from agent 2, i.e., $\pi^1: [\bvec{s}_t, \bvec{i}^2] \mapsto a_t^1$, to optimize its expected cumulative reward:
\begin{align}\label{eq:agent_1_opt}
&\textstyle\max_{\pi^1}~\mathbb{E}\left[\left.\textstyle\sum_{t=0}^{H} r^1_t \right\lvert \bvec{s}_0\right]\\
s.t.\quad
&\bvec{s}_{t+1} \sim p(\bvec{s}_t,a_t), ~a_{t}^1 \sim \pi^1(\bvec{s}_t, \bvec{i}^2), ~ t=0,\ldots,H-1 \nonumber\\
& (s^2_{t})_{t=0}^H \in \bvec{i}^2 \nonumber
\end{align}
where the last constraint means agent 1 knows that agent 2 will comply with its intent. 

Eq. \eqref{eq:agent_2_opt} and \eqref{eq:agent_1_opt} thus provide a formalism for Q1 and Q2.
We next describe our implementation of this formalism for the highway merging scenarios.

%-----------------------
\section{An Implementation for Highway Merging}
\label{sec:implementation}

\begin{figure}[t]\centering
\includegraphics[width=\linewidth]{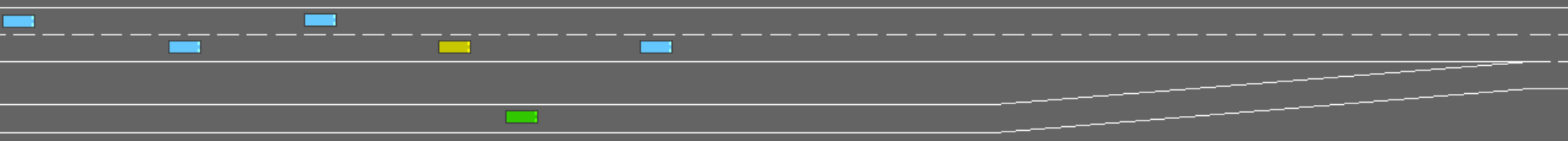}
\caption{Initial configuration of our highway merge environment. The yellow vehicle is the AV capable of sharing its intent. The green vehicle is the merging AV that receives the former's intent. The blue vehicles are human-driven.}
\label{fig:initial_config}
\end{figure}

Under the problem setting formulated in Section \ref{sec:prob_set}, we now conduct a case study on highway merging scenarios to provide an implementation of intent-aware AD and demonstrate its benefits.
The implementation is built upon highway-env \cite{highway-env}, a microscopic highway simulator developed for learning-based AV control. 
Figure \ref{fig:initial_config} shows the initial configuration of the highway merge setup, where the intent-sender AV 2 starts in the rightmost lane of the highway while the intent-receiver AV 1 starts on the entrance ramp. As the merging AV approaches the highway, it has to interact with the intent-sender AV in order to merge successfully. Other vehicles are human-driven.

Specifically, 
1) in Section \ref{sec:Representation of Intents}, we provide a representation for several intents for AV 2, each intent characterizing a reasonable type of its future trajectories;
2) in Section \ref{sec:Complying with the Intents}, we design rule-based policies for AV 2 to comply with its intent, which can viewed as feasible solutions to its policy optimization in Eq. \eqref{eq:agent_2_opt};
and 3) in Section \ref{sec:Representation of Intents}, we propose an RL approach to AV 1's policy optimization in Eq. \eqref{eq:agent_1_opt} that aims to utilize the received intent, with a reward design for safe and efficient merge maneuver.

Our goal with this case study is to demonstrate the benefit of intent-aware AD for the merging AV 1:
knowing the intent of AV 2 on the highway, AV 1 is hypothesized to be able to perform a better merge maneuver than it would without knowing the intent.
We will conduct experiments in Sections \ref{sec:experiments} and \ref{sec:results} to test this hypothesis.

\subsection{Representation of Intents}
\label{sec:Representation of Intents}
As described in Section \ref{sec:prob_set}, an intent encodes information about future behavior of an AV, where the AV that shares its intent also commits to complying with it. 
In this case study, we represent an intent as a restricted set of actions that an intent-sender AV commits to taking for a period of time.
% Specifically, we choose discrete meta-actions to represent the action set of our AV agents as opposed to low-level actions to directly control an agent's acceleration and steering angle. 
In highway-env, we consider the actions to be the following five discrete high-level maneuver decisions (as opposed to low-level controls for acceleration and steering angle), $\mathcal{A} = \{\texttt{IDLE},\allowbreak \texttt{LANE\_LEFT},\allowbreak \texttt{LANE\_RIGHT},\allowbreak \texttt{FASTER},\allowbreak \texttt{SLOWER}\}$. 
Each action corresponds to an update in either the target lane or speed of the AV agent.
Once an action is chosen, the simulator's built-in low-level controllers determine acceleration and steering commands to execute that action. 
% These meta-actions allow us to use maneuver-level representation of vehicles' behavior as our action set.

Then, an intent is specified by a subset of actions, $\mathcal{A}'\subseteq\mathcal{A}$, that the intent-sender AV is restricted to choose from, which can be represented as a vector of indicators:
\begin{align}
\bvec{i}_{\mathcal{A}'} & := \left[\mathbbm{1}_{a\in\mathcal{A}'}\right]_{a \in \mathcal{A}}
\end{align}
where $\mathbbm{1}_E$ is the indicator function of event $E$. 
% Simply put, this means that each element in the intent vector corresponds to an action in the action set $\mathcal{A}$. Value 1 indicates inclusion while 0 indicates exclusion of the corresponding action in the set of committed actions $\mathcal{A}'$. It may be noted that commitment to an intent implies that the agent cannot choose an action outside its committed action set. It also means that each committed action has to be chosen at least once in the time-frame that this intent-commitment is valid.
We also require that each committed action $a\in\mathcal{A}'$ has to be chosen at least once in the committed period of time, since otherwise the intent will reduce to a smaller action subset and cause unnecessary ambiguity.

In this case study, we let the intent-sender AV commit to complying with an intent for the \textit{entire duration} of a simulation episode.

\subsection{AV 2's Compliance with the Intents}
\label{sec:Complying with the Intents}
Since the committed action set $\mathcal{A}'$ is a subset of the original action set $\mathcal{A}$, there are technically $2^5 = 32$ possible intents. However, not all of these possibilities may be feasible or meaningful. For instance, an empty $\mathcal{A}'$ is not feasible because no action (\texttt{IDLE}) is also an action in our action set. Further, when only one action is committed to, it may not be feasible for the AV to keep choosing \texttt{LANE\_LEFT}, \texttt{LANE\_RIGHT}, \texttt{FASTER} or \texttt{SLOWER}. 
% This is especially the case when the commitment duration $H$ is sufficiently larger than the interval at which action-decisions are made. 
Hence, in our setup, we always include \texttt{IDLE} action in $\mathcal{A}'$.
When we scale up to two committed actions, we see four possibilities. 
Out of these, $\mathcal{A}'=\{\texttt{IDLE},\allowbreak \texttt{LANE\_RIGHT}\}$ is not feasible in our case because the intent-sender AV is already on the rightmost lane of the highway. 
% The committed action set can be scaled up further in a similar fashion. 
This results in the following four intents that we will consider:
\begin{align*}
    \bvec{i}_{\texttt{IDLE}} & = [1,\,0,\,0,\,0,\,0] \\
    \bvec{i}_{\texttt{LANE\_LEFT}} & = [1,\,1,\,0,\,0,\,0] \\
    \bvec{i}_{\texttt{FASTER}} & = [1,\,0,\,0,\,1,\,0] \\
    \bvec{i}_{\texttt{SLOWER}} & = [1,\,0,\,0,\,0,\,1]
    .
\end{align*}
% $\bvec{i}_{\texttt{IDLE}}$ or idle intent means that the intent-sender AV is only allowed to choose the \texttt{IDLE} action during the course of its commitment. Similarly, $\bvec{i}_{\texttt{LANE\_LEFT}}$ or lane-changing (LC) intent implies that only actions \texttt{IDLE} and \texttt{LANE\_LEFT} can be chosen. One can intuitively interpret idle intent as a non-cooperative intent where the mainstream AV chooses to stick to its current trajectory without making changes to accommodate the merging vehicle. Likewise, LC intent can be seen as a cooperative intent wherein the mainstream vehicle commits to a lane change to the left to create a gap for the merging vehicle.

We manually design rule-based policies for the intent-sender AV 2 to comply with one of the intents. 
First, we set its initial speed to $30$ m/s and assign it an intent randomly sampled from a uniform distribution on $\{\bvec{i}_{\texttt{IDLE}},\allowbreak \bvec{i}_{\texttt{LANE\_LEFT}},\allowbreak \bvec{i}_{\texttt{FASTER}},\allowbreak \bvec{i}_{\texttt{SLOWER}}\}$.
Then AV 2 acts according to the behavior policy specific to its intent.
% We also assume that this intent commitment is for the entire duration of a simulation episode. 
For intent $\bvec{i}_{\texttt{IDLE}}$, the AV 2 just chooses \texttt{IDLE} throughout an episode. 
For other intents, it also needs to decide when the other action will be taken. 
We facilitate this decision by assigning an action-trigger position for the AV to take the committed action. 
Concretely, AV 2 chooses \texttt{IDLE} till it reaches the assigned action-trigger position, where it chooses the committed action, namely \texttt{LANE\_LEFT}, \texttt{FASTER}, or \texttt{SLOWER}. Thereafter, it continues to choose \texttt{IDLE}.
The action-trigger position is randomly chosen from three options along AV 2's trajectory before the end of the merging zone, which introduces diversity in admissible future trajectories.

% comment on different action-trigger positions.

% This behavior policy for the mainstream vehicle allows us to design a minimal working setup to test the usefulness of intent-sharing. 
 
\subsection{AV 1's RL-based Intent Utilization}
\label{sec:RL-based Intent Utilization}
We now describe our RL-based approach to AV 1's utilization of the received intent, for which we specify the state structure, and reward function below.

\subsubsection{State}
AV 1 conditions its policy on the world state $\bvec{s}$ and AV 2's intent $\bvec{i}^2$, where the world state is further decomposed into AV 1's local state $\bvec{s}^1$, AV 2's local state $\bvec{s}^2$, and the external state $\bvec{s}^e$ (Section \ref{sec:prob_set}). 
In our case study, human-driven vehicles $\mathcal{K}$ constitute the external state. 

To implement this, we use vehicles' kinematics to represent their state vectors:
\begin{align}
    \bvec{s}^e & = \left[[x^k,\, y^k,\, v_x^k,\, v_y^k]\right]_{k \in \mathcal{K}}\\
    \bvec{s}^1 & = [x^1,\, y^1,\, v_x^1,\, v_y^1]\\
    \bvec{s}^2 & = [x^2,\, y^2,\, v_x^2,\, v_y^2]
\end{align}
where $x^u$ and $y^u$ are $x$ and $y$ positions of vehicle $u$, and $v_x^u$ and $v_y^u$ are $x$ and $y$ components of vehicle $u$'s velocity, respectively.

To interpret and use the received intent information from AV 2, we consider an auxiliary state due to communication:
\begin{align}
\bvec{z}^{1\leftarrow 2} & = 
    \begin{cases}
        \bvec{i}^2, & \text{if AV 2 shares its intent with AV 1} \\
        \bvec{0}, & \text{otherwise.}
    \end{cases}
    % \bvec{z}^{1\leftarrow 2} = \bvec{i}^2
% \bvec{z}^{1\leftarrow k} & = \bvec{0}, \forall \, k \neq 2
\end{align} 

% We set $\bvec{z}^{1\leftarrow e} = \bvec{0}$ for $k \neq 2$ since other vehicles are not capable of intent-sharing. 
AV 1's policy is now conditioned on the world state $\bvec{s}$ and the auxiliary state $\bvec{z}^{1\leftarrow 2}$.
The introduction of an auxiliary state allows us to also investigate cases when AV 2 does not share its intent, as we will discuss in Section \ref{sec:experiments}.

\subsubsection{Reward}
Our reward structure for AV 1 is composed of four components:
\begin{align}
    r_t^1 & = r_t^s + r_t^l + r_t^c + r_t^m
    .
\end{align}

Component $r_t^s$ encourages high speed: 
\begin{align*}
    r_t^s & = \beta^s \frac{v_t^1 - v_{\min}^1}{v_{\max}^1 - v_{\min}^1}
\end{align*}
where $v_t^1$, $v_{\min}^1$, and $v_{\max}^1$ are respectively the current, minimum, and maximum speeds of  AV 1. We set $v_{\min}^1 = 20$ m/s, and $v_{\max}^1 = 30$ m/s. 

Component $r_t^l$ encourages AV 1 to seek and stay in the rightmost lane of the highway:
\begin{align*}
    r_t^l & = 
    \begin{cases}
        \beta^l, & \text{if AV 1 is in the rightmost lane} \\
        % & \text{of the highway} \\
        0, & \text{otherwise.}
    \end{cases}
\end{align*}

Component $r_t^c$ penalizes collision with other vehicles and road objects:
\begin{align*}
    r_t^c & = 
    \begin{cases}
        \beta^c, & \text{if AV 1 has crashed} \\
        0, & \text{otherwise.}
    \end{cases}
\end{align*}

Component $r_t^m$ is the reward that AV 1 obtains when it merges successfully, which is further decomposed into four terms:
\begin{align*}
    r_t^m & = 
    \begin{cases}
        r^q + r^{f} + r^{r} + r^e, & \text{if successful merge at $t$} \\
        0, & \text{otherwise.}
    \end{cases}
\end{align*}
These terms scale the merge reward per the quality of merge maneuver:
\begin{itemize}
\item 
Term $r^q$ encourages quick merging via
$
r^q = \beta^q \frac{1}{t_m}
$,
where $t_m$ is the time at which merge occurred.
\item 
Terms $r^{f}$ and $r^{r}$ are gap-based rewards that encourage safe merging: % by penalizing unsafe gaps with the front and rear vehicles at merge:
$    r^{f}  = \beta^{f} \min \left\{\log \frac{\Delta d_m^{f,1}}{t_{*}^h \,v_m^1},\, 0 \right\}
$
and
$
    r^{r}  = \beta^{r} \min \left\{\log \frac{\Delta d_m^{1,r}}{t_{*}^h \,v_m^r},\, 0 \right\}
$.
Here, $r^{f}$ and $r^{r}$ correspond to gaps with the front ($f$) and rear ($r$) vehicles respectively. $\Delta d_m^{f,1}$ and $\Delta d_m^{1,r}$ are relative distances of AV 1 with vehicles $f$ and $r$, respectively, at merge-time $t_m$. $t_{*}^h$ is the desired time-headway, which we set to $1.2$ s. $v_m^1$ and $v_m^r$ are speeds of AV 1 and vehicle $r$ at $t_m$. The structure of these rewards is based on \cite{liu2022autonomous} and seeks to aggressively penalize AV 1 when it merges with a time-headway less than $t_{*}^h$. If there is no vehicle in front of or behind AV 1, these rewards are set to $0$.
\item 
Term $r^e$ is the component of $r_t^m$ that rewards AV 1 for an efficient merge, by penalizing it for merging with speeds deviating from its target speed:
$
    r^e = \beta^e \left| \frac{v_\text{*}^1 - v_m^1}{v_{*}^1} \right|
$,
where $v_{*}^1$ and $v_m^1$ are AV 1's target speed and speed at $t_m$, respectively. We set $v_{*}^1 = 30$ m/s.
\end{itemize}

$\beta$'s in the above equations are scaling coefficients to scale their corresponding rewards. We set $\beta^s = 0.275$, $\beta^l = 0.1$, $\beta^c = -5.0$, $\beta^q = 2.0$, $\beta^{f} = 0.5$, $\beta^{r} = 0.5$, and $\beta^e = -1.0$ after manual tuning so as to encourage reasonable and safe merging behavior.

We set the initial speed of AV $1$ to $20$ m/s.

%-----------------------
\section{Experimental Setup} % Experiments
\label{sec:experiments}

We design our environment with one intent-sender AV (AV 2) and four human-driven vehicles on the highway, and one merging AV (AV 1) on the entrance ramp (Figure \ref{fig:initial_config}). We use human-driven vehicles in our experiments to create a realistic traffic scenario for the merging AV to learn from. However, these vehicles are incapable of intent-sharing, and their behavior is executed as a part of the environment. In highway-env, the behavior of a human driver is modeled using the Intelligent Driver Model (IDM) \cite{treiber2000congested} for longitudinal motion and Minimizing Overall Braking Induced by Lane change (MOBIL) model \cite{kesting2007general} for lateral motion.
% Further, the initial speeds of the merging AV and the mainstream AV are set to $20$ m/s and $30$ m/s, respectively.

\subsection{Scenarios}
Intuitively, the merging AV should be able to use extra information about the intent-sender AV (which we also call mainstream AV) to better characterize its state, which should in turn help it make better decisions about merging. To test this conjecture, we consider two scenarios for our experiments: with and without intent-sharing. In the first case, the merging AV can receive the intent of the mainstream intent-sender AV. In the second case, however, intent-sharing is blocked, and the merging AV does not get information about the mainstream AV's intent. However, it is important to note that the mainstream AV still behaves as per its intent in both the scenarios, irrespective of whether it shares that intent or not.
For both scenarios, we randomly choose an intent out of the four candidates to start a training episode.

% Further, the initial speeds of the merging AV and the mainstream AV are set to $20$ m/s and $30$ m/s, respectively.

\begin{table*}[t] % !h, !ht, !htbp
\begin{center}
\caption{\label{tab:performance}
Performance of the Learned Merging Policies with and without Intent-Sharing.}
% \begin{tabular}{|c|c|c|c|c|c|}
\begin{tabular}{ c c c c c c }
    %\hline
    %\multicolumn{2}{|c|}{Without Intent-Sharing} \\
    \hline
    %\multicolumn{2}{|c|}{Idle Intent} & \multicolumn{2}{|c|}{LC Intent}\\
        % \multirow{2}{*}{Intent} & \multirow{2}{*}{Action-Trigger Positions (m)} & \multicolumn{2}{c|}{With Intent-Sharing} & \multicolumn{2}{c|}{Without Intent-Sharing}\\
        \multirow{2}{*}{Intent} & \multirow{2}{*}{Action-Trigger Positions (m)} & \multicolumn{2}{c}{With Intent-Sharing} & \multicolumn{2}{c}{Without Intent-Sharing}\\
        
    % \cline{3-6}
    & & Cumulative Reward & Crash Rate (\%) & Cumulative Reward & Crash Rate (\%) \\
    \hline%\hline
    $\bvec{i}_{\texttt{IDLE}}$ & N/A & $\mathbf{2.725 \pm 0.087}$ & $\mathbf{0.0}$ & $1.416 \pm 2.566$ & $20.0$ \\
    \hline
    \multirow{3}{*}{$\bvec{i}_{\texttt{LANE\_LEFT}}$} & $220$ & $\mathbf{3.616 \pm 0.000}$ & $\mathbf{0.0}$ & $3.144 \pm 0.338$ & $\mathbf{0.0}$ \\
    % \cline{2-6}
    & $250$ & $\mathbf{3.481 \pm 0.000}$ & $\mathbf{0.0}$ & $2.855 \pm 0.577$ & $\mathbf{0.0}$ \\
    % \cline{2-6}
    & $280$ & $\mathbf{3.481 \pm 0.000}$ & $\mathbf{0.0}$ & $2.701 \pm 0.514$ & $\mathbf{0.0}$ \\
    \hline
    \multirow{3}{*}{$\bvec{i}_{\texttt{FASTER}}$} & $190$ & $\mathbf{3.149 \pm 0.000}$ & $\mathbf{0.0}$ & $2.848 \pm 0.389$ & $\mathbf{0.0}$ \\
    % \cline{2-6}
    & $220$ & $\mathbf{2.993 \pm 0.000}$ & $\mathbf{0.0}$ & $2.660 \pm 0.354$ & $\mathbf{0.0}$ \\
    % \cline{2-6}
    & $250$ & $\mathbf{2.767 \pm 0.000}$ & $\mathbf{0.0}$ & $2.587 \pm 0.272$ & $\mathbf{0.0}$ \\
    \hline
    \multirow{3}{*}{$\bvec{i}_{\texttt{SLOWER}}$} & $160$ & $\mathbf{2.957 \pm 0.742}$ & $\mathbf{0.0}$ & $1.490 \pm 3.072$ & $20.0$ \\
    % \cline{2-6}
    & $190$ & $\mathbf{2.898 \pm 0.562}$ & $\mathbf{0.0}$ & $1.436 \pm 2.894$ & $20.0$ \\
    % \cline{2-6}
    & $220$ & $\mathbf{2.736 \pm 0.486}$ & $\mathbf{0.0}$ & $-0.121 \pm 3.010$ & $40.0$ \\
    \hline
\end{tabular}
\end{center}
\end{table*}

\begin{figure*}[ht]
\centering
% 0000
    \begin{subfigure}[b]{0.42\linewidth} 
         \centering
         \includegraphics[width=\textwidth, trim={2.04cm 4.67cm 1.6cm 4.6cm}, clip]{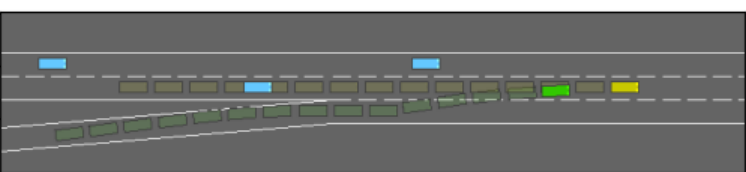}
         \caption{\texttt{IDLE} -- With Intent-Sharing}
         \label{fig:w_0000}
     \end{subfigure}
     \qquad
     \begin{subfigure}[b]{0.42\linewidth}
         \centering
         \includegraphics[width=\textwidth, trim={2.04cm 4.67cm 1.6cm 4.6cm}, clip]{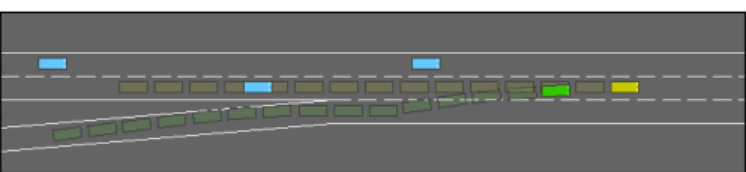}
         \caption{\texttt{IDLE} -- Without Intent-Sharing}
         \label{fig:wo_0000}
     \end{subfigure}
% 1000 - 280
     \begin{subfigure}[b]{0.42\linewidth}
         \centering
         \includegraphics[width=\textwidth, trim={2.04cm 4.67cm 1.6cm 4.6cm}, clip]{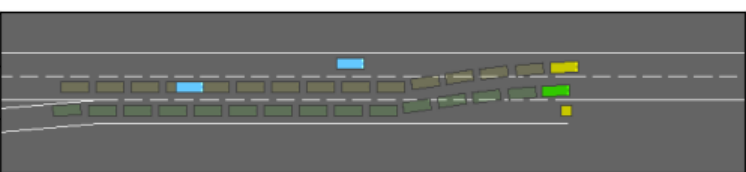}
         \caption{\texttt{LANE\_LEFT} at 280 m -- With Intent-Sharing}
         \label{fig:w_1000_280}
     \end{subfigure}
     \qquad
     \begin{subfigure}[b]{0.42\linewidth}
         \centering
         \includegraphics[width=\textwidth, trim={2.04cm 4.67cm 1.6cm 4.6cm}, clip]{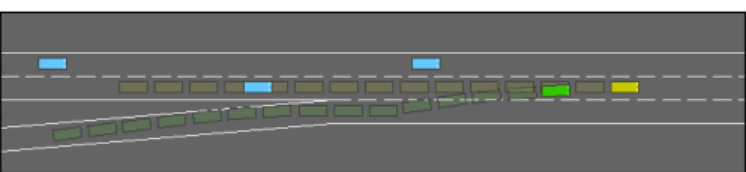}
         \caption{\texttt{LANE\_LEFT} at 280 m -- Without Intent-Sharing}
         \label{fig:wo_1000_280}
     \end{subfigure}
% 0010 - 220
     \begin{subfigure}[b]{0.42\linewidth}
         \centering
         \includegraphics[width=\textwidth, trim={2.04cm 4.67cm 1.6cm 4.6cm}, clip]{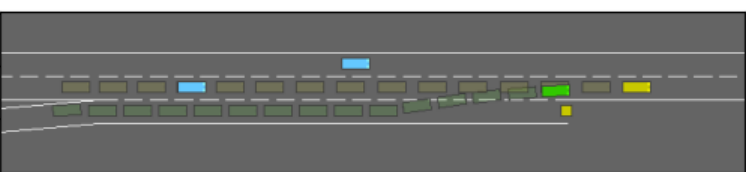}
         \caption{\texttt{FASTER} at 220 m  -- With Intent-Sharing}
         \label{fig:w_0010_220}
     \end{subfigure}
     \qquad
     \begin{subfigure}[b]{0.42\linewidth}
         \centering
         \includegraphics[width=\textwidth, trim={2.04cm 4.67cm 1.6cm 4.6cm}, clip]{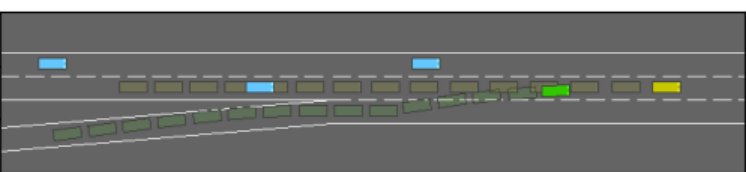}
         \caption{\texttt{FASTER} at 220 m  -- Without Intent-Sharing}
         \label{fig:wo_0010_220}
     \end{subfigure}
% 0001 - 190
     \begin{subfigure}[b]{0.42\linewidth}
         \centering
         \includegraphics[width=\textwidth, trim={2.04cm 4.67cm 1.6cm 4.6cm}, clip]{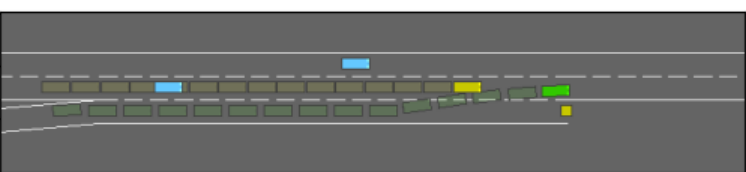}
         \caption{\texttt{SLOWER} at 190 m -- With Intent-Sharing}
         \label{fig:w_0001_190}
     \end{subfigure}
     \qquad
     \begin{subfigure}[b]{0.42\linewidth}
         \centering
         \includegraphics[width=\textwidth, trim={2.04cm 4.67cm 1.6cm 4.6cm}, clip]{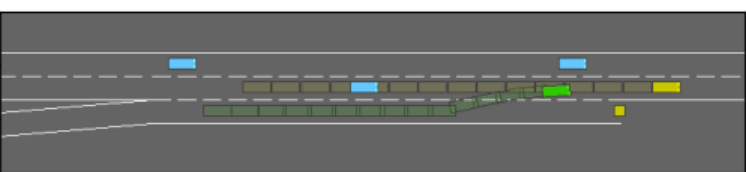}
         \caption{\texttt{SLOWER} at 190 m -- Without Intent-Sharing}
         \label{fig:wo_0001_190}
     \end{subfigure}
\caption{Snapshots at merge with visible past trajectories.
For $\bvec{i}_{\texttt{IDLE}}$, the merging AV learns the same policy both with and without intent-sharing (\ref{fig:w_0000}, \ref{fig:wo_0000}). It positions itself to merge behind the mainstream AV, whose intent indicates that it is going to stick to its current trajectory without making changes to accommodate the merging AV.
For $\bvec{i}_{\texttt{LANE\_LEFT}}$, Figure \ref{fig:w_1000_280} shows that based on its training, the merging AV has learned that the mainstream AV will definitely change its lane by 280 m. So it triggers a lane change at the same time-step. Without intent-sharing, however, the merging AV doesn’t know about the lane change, so it positions itself behind the mainstream AV and does not wait for it to change lanes (\ref{fig:wo_1000_280}). 
For $\bvec{i}_{\texttt{FASTER}}$, when intent is shared, the merging AV merges towards the end of the merging zone, in anticipation that the mainstream AV will speed up (\ref{fig:w_0010_220}). On the other hand, without intent-sharing, the merging AV just positions itself behind the mainstream AV irrespective of what the mainstream AV does (\ref{fig:wo_0010_220}).
For $\bvec{i}_{\texttt{SLOWER}}$, when the merging AV knows the mainstream AV's intent, it is able to merge in front of this slow-moving AV while moving at its target speed (\ref{fig:w_0001_190}). However, when it does not receive the latter's intent, it is forced to slow down and merge behind the already slow-moving mainstream AV (\ref{fig:wo_0001_190}).
}
\label{fig:traj}
\end{figure*}

\subsection{RL Algorithm Details}
While our problem setting consists of multiple agents, we have pre-decided policies for our intent-sender AV. Therefore, we can use single-agent RL methods to learn an optimal behavior policy for the merging AV. Specifically, we use Stable Baselines3's \cite{stable-baselines3} implementation of Deep Q Network (DQN) \cite{mnih2013playing} using multi-layer perceptrons (MLP) for both the scenarios of our experiments. We use two hidden layers with 512 neurons each and adjust the following hyperparameters: \texttt{learning\_rate} = 5e-4, \texttt{buffer\_size} = 15000, \texttt{learning\_starts} = 1000, \texttt{batch\_size} = 32, \texttt{gamma} = 0.95, \texttt{train\_freq} = 1, \texttt{gradient\_steps} = 1, and \texttt{target\_update\_interval} = 50. We learn for 40,000 steps for each of  five different seeds.
% Algo - DQN

%-----------------------
\section{Results and Discussion}
\label{sec:results}

Table \ref{tab:performance} shows the performance of policies learned by the merging AV. First, we report the average cumulative reward that the merging AV receives by the end of a simulation episode. We also report the standard error. Second, we report the percentage of cases (seeds) in which the merging AV is unable to learn a crash-free policy. We also decompose performance with respect to the four intent cases and their respective action-trigger positions. It may be noted that action-trigger positions are set as: (a) 220 m, 250 m, and 280 m for $\bvec{i}_{\texttt{LANE\_LEFT}}$, (b) 190 m, 220 m, and 250 m for $\bvec{i}_{\texttt{FASTER}}$, and (c) 160 m, 190 m, and 220 m for $\bvec{i}_{\texttt{SLOWER}}$. These values correspond to cases for early, intermediate and late intent-realization. For context, the merging zone in our setup extends from 230 m to 310 m. Further, action-trigger positions vary across intents because we observed that not all positions of intent-realization allow sufficient time for the merging AV to utilize that intent, and that these positions vary across different intents.

First of all, we see that the merging AV is able to get higher cumulative rewards when intent-sharing is on as compared to when it is turned off (better values are in bold). The variation in rewards is also significantly lower in the intent-sharing case, especially with $\bvec{i}_{\texttt{LANE\_LEFT}}$ and $\bvec{i}_{\texttt{FASTER}}$ where the same merging policy is learned for all the seeds used in our experiments. On the other hand, merging policies without intent-sharing show high variation in cumulative rewards. This makes sense intuitively because the merging AV is uncertain about the behavior of the intent-sender mainstream AV and thus, tries to learn a general policy if it is unaware of the mainstream AV's intent. Another point to note is that the policies learned with intent-sharing are crash-free, while crashes are observed for some seeds in the $\bvec{i}_{\texttt{IDLE}}$ and $\bvec{i}_{\texttt{SLOWER}}$ cases without intent-sharing.

A closer look at the cumulative rewards obtained for different intents reveals more interesting trends. We see that, for the initial configuration used in our experiments, the merging AV performs best when the mainstream AV follows $\bvec{i}_{\texttt{LANE\_LEFT}}$. Furthermore, even within a given intent, the position at which intent-related actions are triggered influence the cumulative rewards that the merging AV is able to get. This indicates that the mainstream AV also has a choice to make -- the intent it should pursue and the sequence of actions it should take while complying with that intent. This decision can be influenced by multiple factors such as the configuration of vehicles on the highway and the entrance ramp, and the relative importance that the intent-sender AV gives to the merging AV's interests as compared to its own.

% %----------------------- 

% \addtolength{\textheight}{-5cm}   % This command serves to balance the column lengths
%                                   % the textheight of the last page by a suitable amount.
%                                   % This command does not take effect until the next page
%                                   % so it should come on the page before the last. Make
%                                   % sure that you do not shorten the textheight too much.

For visual inspection of the learned merging behavior, Figure \ref{fig:traj} provides snapshots of the rendered environment at the instant the merging AV successfully merges, using policies from one set of our learned models. The snapshots also show past positions of the intent-sender mainstream AV and the merging AV using transparent (faded) yellow and green colors respectively. We see that, with intent-sharing, the merging AV is able to learn different policies for each intent. However, in the absence of intent-sharing, it tries to learn a general policy while trying to handle uncertainty in the mainstream AV's behavior. We further discuss individual intent cases in detail in the caption.

\section{Conclusion and Future Work} 
\label{sec:conclusion}

In this work, we formulate intent-aware AD as a multi-agent decision problem. Using a case study with two AVs in a highway merging scenario, we further demonstrate that sharing of intent facilitates learning of robust and crash-free behavior policies for the intent-receiver AV agent by effectively adapting to the intended behavior of an intent-sender AV. 
% We also make the case for choosing an optimal intent and an optimal behavior policy to realize that intent. 
% Therefore, as our next step, we will expand our implementation to allow the intent-sender agent to learn its optimal behavior, in terms of the intent it should share and the actions it should execute in compliance with that intent.
% This work can be extended along several possible directions.
% Several directions can be pursued as future work.
As future work, several directions can be explored.
% First, our current implementation only learns the behavior of the agent that receives intent. 
A reasonable next step is then to extend our framework to allow choosing of intent via joint learning of the intent-sender and intent-receiver agents.
Another direction is to scale beyond the two-agent-one-intent setting and allow communication of intents from multiple intent-sender agents to multiple intent-receiver agents.
It is also possible to extend to scenarios beyond merging.
% (e.g., other scenarios in highway-env such as the roundabout and intersection environments).
% Mixed-autonomy

%-----------------------
% %----------------------- 

%                                   % the textheight of the last page by a suitable amount.
%                                   % This command does not take effect until the next page
%                                   % so it should come on the page before the last. Make
%                                   % sure that you do not shorten the textheight too much.

%%%%%%%%%%%%%%%%%%%%%%%%%%%%%%%%%%%%%%%%%%%%%%%%%%%%%%%%%%%%%%%%%%%%%%%%%%%%%%%%

%%%%%%%%%%%%%%%%%%%%%%%%%%%%%%%%%%%%%%%%%%%%%%%%%%%%%%%%%%%%%%%%%%%%%%%%%%%%%%%%



\section*{Acknowledgment}
This work is supported in part by NSF IIS-2154904, CNS-2213731.

\bibliographystyle{IEEEtran.bst}
\bibliography{references}

\begin{thebibliography}{10}
\providecommand{\url}[1]{#1}
\csname url@rmstyle\endcsname
\providecommand{\newblock}{\relax}
\providecommand{\bibinfo}[2]{#2}
\providecommand\BIBentrySTDinterwordspacing{\spaceskip=0pt\relax}
\providecommand\BIBentryALTinterwordstretchfactor{4}
\providecommand\BIBentryALTinterwordspacing{\spaceskip=\fontdimen2\font plus
\BIBentryALTinterwordstretchfactor\fontdimen3\font minus
  \fontdimen4\font\relax}
\providecommand\BIBforeignlanguage[2]{{%
\expandafter\ifx\csname l@#1\endcsname\relax
\typeout{** WARNING: IEEEtran.bst: No hyphenation pattern has been}%
\typeout{** loaded for the language `#1'. Using the pattern for}%
\typeout{** the default language instead.}%
\else
\language=\csname l@#1\endcsname
\fi
#2}}

\bibitem{on2021taxonomy}
O.-R. A.~D. Committee, ``Taxonomy and definitions for terms related to
  cooperative driving automation for on-road motor vehicles,'' SAE
  International, Tech. Rep., 2021.

\bibitem{highway-env}
E.~Leurent, ``An environment for autonomous driving decision-making,''
  \url{https://github.com/eleurent/highway-env}, 2018.

\bibitem{bandyopadhyay2013intention}
T.~Bandyopadhyay, K.~S. Won, E.~Frazzoli, D.~Hsu, W.~S. Lee, and D.~Rus,
  ``Intention-aware motion planning,'' in \emph{Algorithmic Foundations of
  Robotics X: Proceedings of the Tenth Workshop on the Algorithmic Foundations
  of Robotics}.\hskip 1em plus 0.5em minus 0.4em\relax Springer, 2013, pp.
  475--491.

\bibitem{qi2018intent}
S.~Qi and S.-C. Zhu, ``Intent-aware multi-agent reinforcement learning,'' in
  \emph{2018 IEEE international conference on robotics and automation
  (ICRA)}.\hskip 1em plus 0.5em minus 0.4em\relax IEEE, 2018, pp. 7533--7540.

\bibitem{wu2021spatial}
J.~Wu, X.~Sun, A.~Zeng, S.~Song, S.~Rusinkiewicz, and T.~Funkhouser, ``Spatial
  intention maps for multi-agent mobile manipulation,'' in \emph{2021 IEEE
  International Conference on Robotics and Automation (ICRA)}.\hskip 1em plus
  0.5em minus 0.4em\relax IEEE, 2021, pp. 8749--8756.

\bibitem{kim2021communication}
W.~Kim, J.~Park, and Y.~Sung, ``Communication in multi-agent reinforcement
  learning: Intention sharing,'' in \emph{International Conference on Learning
  Representations}, 2021.

\bibitem{wang2022multi}
H.~M. Wang, S.~S. Avedisov, O.~Altintas, and G.~Orosz, ``Multi-vehicle conflict
  management with status and intent sharing,'' in \emph{2022 IEEE Intelligent
  Vehicles Symposium (IV)}.\hskip 1em plus 0.5em minus 0.4em\relax IEEE, 2022,
  pp. 1321--1326.

\bibitem{matthews2017intent}
M.~Matthews, G.~Chowdhary, and E.~Kieson, ``Intent communication between
  autonomous vehicles and pedestrians,'' \emph{arXiv preprint
  arXiv:1708.07123}, 2017.

\bibitem{tang2019towards}
Y.~Tang, ``Towards learning multi-agent negotiations via self-play,'' in
  \emph{Proceedings of the IEEE/CVF International Conference on Computer Vision
  Workshops}, 2019, pp. 0--0.

\bibitem{liu2022autonomous}
Q.~Liu, F.~Dang, X.~Wang, and X.~Ren, ``Autonomous highway merging in mixed
  traffic using reinforcement learning and motion predictive safety
  controller,'' in \emph{2022 IEEE 25th International Conference on Intelligent
  Transportation Systems (ITSC)}.\hskip 1em plus 0.5em minus 0.4em\relax IEEE,
  2022, pp. 1063--1069.

\bibitem{triest2020learning}
S.~Triest, A.~Villaflor, and J.~M. Dolan, ``Learning highway ramp merging via
  reinforcement learning with temporally-extended actions,'' in \emph{2020 IEEE
  Intelligent Vehicles Symposium (IV)}.\hskip 1em plus 0.5em minus 0.4em\relax
  IEEE, 2020, pp. 1595--1600.

\bibitem{bouton2019cooperation}
M.~Bouton, A.~Nakhaei, K.~Fujimura, and M.~J. Kochenderfer, ``Cooperation-aware
  reinforcement learning for merging in dense traffic,'' in \emph{2019 IEEE
  Intelligent Transportation Systems Conference (ITSC)}.\hskip 1em plus 0.5em
  minus 0.4em\relax IEEE, 2019, pp. 3441--3447.

\bibitem{toghi2022social}
B.~Toghi, R.~Valiente, D.~Sadigh, R.~Pedarsani, and Y.~P. Fallah, ``Social
  coordination and altruism in autonomous driving,'' \emph{IEEE Transactions on
  Intelligent Transportation Systems}, vol.~23, no.~12, pp. 24\,791--24\,804,
  2022.

\bibitem{treiber2000congested}
M.~Treiber, A.~Hennecke, and D.~Helbing, ``Congested traffic states in
  empirical observations and microscopic simulations,'' \emph{Physical review
  E}, vol.~62, no.~2, p. 1805, 2000.

\bibitem{kesting2007general}
A.~Kesting, M.~Treiber, and D.~Helbing, ``General lane-changing model mobil for
  car-following models,'' \emph{Transportation Research Record}, vol. 1999,
  no.~1, pp. 86--94, 2007.

\bibitem{stable-baselines3}
\BIBentryALTinterwordspacing
A.~Raffin, A.~Hill, A.~Gleave, A.~Kanervisto, M.~Ernestus, and N.~Dormann,
  ``Stable-baselines3: Reliable reinforcement learning implementations,''
  \emph{Journal of Machine Learning Research}, vol.~22, no. 268, pp. 1--8,
  2021. [Online]. Available: \url{http://jmlr.org/papers/v22/20-1364.html}
\BIBentrySTDinterwordspacing

\bibitem{mnih2013playing}
V.~Mnih, K.~Kavukcuoglu, D.~Silver, A.~Graves, I.~Antonoglou, D.~Wierstra, and
  M.~Riedmiller, ``Playing atari with deep reinforcement learning,'' 2013.

\end{thebibliography}

\end{document}